%% file: iclr2025_conference.tex
\documentclass{article} % For LaTeX2e
\usepackage{iclr2025_conference,times}

\input{math_commands.tex}

\usepackage{subcaption}

\usepackage{hyperref}
\usepackage{url}
\usepackage[ruled,vlined]{algorithm2e}
\usepackage{amsmath}
\usepackage{mathtools}
\usepackage{booktabs}
\usepackage{multirow}
\usepackage{tikz-cd}
\usepackage{rotating}
\usepackage{pdflscape}

\usepackage{amssymb}

\title{Amortized Thompson Sampling and Efficient Information-Directed Sampling for Simulator-Based Design}

\author{Michal Kobiela, Michael U. Gutmann \\
School of Informatics\\
The University of Edinburgh \\
Edinburgh, UK \\
\texttt{\{s1870794, michael.gutmann\}@ed.ac.uk}
\AND
}

\iclrfinalcopy 
\begin{document}

\maketitle

\begin{abstract}
Thompson sampling is a widely used strategy for sequential decision-making under uncertainty, offering a principled way to balance exploration and exploitation. This strategy involves maximizing a target objective under a single sample drawn from the current belief over unknown quantities. Conventionally, this optimization is performed once the belief has been inferred from data, and it becomes computationally burdensome when many Thompson samples are required, especially in continuous action spaces. In this work, we introduce an approach that amortizes this process ahead of time, before any data is collected, so that an approximation of the full Thompson sampling distribution is available immediately once observations arrive. We employ this approach to solve the multi-round setting, involving sequential data acquisition and Thompson-sample recommendation across rounds. Beyond accelerating Thompson sampling, our approach enables efficient estimation of expected information gain about the optimal action. This can be used to select more informative Thompson samples or to obtain maximally informative designs with optimization. Combining this estimate of expected information gain with immediate regret supports information-directed sampling (IDS), which can outperform Thompson sampling by explicitly balancing these two quantities. We demonstrate our approach on two simulator-based design problems in synthetic biology, where the simulators have intractable likelihoods.
\end{abstract}

\section{Introduction}

Thompson sampling is a widely used strategy for sequential decision-making under uncertainty. Given the observations collected so far, it samples a plausible value of the unknown quantities from their posterior distribution and selects an action that is optimal under that sample. This provides a principled mechanism for balancing exploration and exploitation and has strong theoretical guarantees under established assumptions. The strategy was first proposed in 1933 by \citet{thompson1933}, and later analyzed by \citet{russo2014learning}, who provided Bayesian regret bounds  for this strategy. The method proved to be effective as decision making principle for multi-armed bandits \citep{chapelle2011empirical} and Bayesian optimization \citep{eriksson2019scalable,garnett}. However, conventional Thompson sampling can be burdensome in continuous action domains because each posterior sample requires solving a separate optimization problem.

To illustrate how this computational cost recurs, consider a laboratory using a simulator to design experiments.  First, generating multiple Thompson samples within a single experimental round requires solving a separate optimization problem for each posterior sample. Multiple samples may be needed to obtain a batch of recommendations or to characterize uncertainty over the optimal action, see \citep[see e.g.][]{kobiela2026risk}. Second, after each experiment, the posterior changes as new data become available, requiring inference and optimization to be performed again. When delays between experiments incur costs, the next action should be available with minimal latency. Third, when the method is deployed under different conditions, for example, in another laboratory with different unknown experimental conditions, these steps must be repeated using observations from the new setting. These repeated computations are particularly expensive when the objective is evaluated using a costly, stochastic, or non-differentiable simulator with continuous action space.

In this work, we amortize the optimization underlying Thompson sampling before data collection. Our approach trains a parameterized policy to map values of the unknown model parameters to approximately reward-maximizing actions under those parameter values. We train this policy using policy gradient method, allowing the approach to handle stochastic and non-differentiable simulators. The trained policy is then combined with simulation-based inference (SBI): once experimental observations become available, SBI provides samples from the posterior over the unknown parameters, and passing these samples through the policy produces approximate Thompson samples. In this way, the costly optimization is performed during offline training rather than repeatedly for each posterior sample and experimental round.

We also develop a second amortized sampling route that bypasses explicit parameter inference. We train an \emph{action flow}, implemented as a conditional normalizing flow, to represent a conditional distribution over approximate Thompson samples given an experimental history. Training targets are generated by applying the pretrained policy to parameter values sampled from the prior. Both approaches support sequential design through round-specific flows trained offline over a prescribed experimental horizon and conditioned at deployment on the evolving history of actions and observations. The pretrained models can be reused across experimental settings whose unknown conditions are represented by the simulator and training prior.

Efficient access to an approximate distribution over optimal actions, together with amortization across design rounds, provides capabilities beyond accelerating conventional Thompson sampling. Standard Thompson sampling balances exploration and exploitation implicitly through posterior randomization, without explicitly evaluating the information value of a recommendation. To enable more directed exploration, the key observation is that expected information gain about the optimal action is the expected reduction in the entropy of the Thompson-sampling distribution after observing the outcome of an experiment from predictive distribution. The tractable conditional density of the action flow enables efficient estimation of this quantity, allowing us to score and select Thompson-sampled candidates to avoid uninformative measurements and focus experimental effort on identifying the optimal action. The parameter flows additionally support one-step look-ahead regret estimation by conditioning on hypothetical observations. This also encourages exploration as the immediate action outcome is unimportant and only expected outcome of the future Thompson sample is used.

In contrast, selecting actions by minimizing expected immediate regret favors exploitation: these actions maximize expected reward under the current belief but may provide little information for future rounds. Together, these criteria allow explicit control over the exploration–exploitation trade-off beyond posterior randomization. For example, in the final round of experimental design, an exploitative action may be preferable, whereas earlier rounds may benefit from informative experiments that help identify the optimal action, but not necessarily yield good reward. Neither purely exploitative nor purely information-seeking actions are necessarily preferable throughout an experiment: exploitation can leave useful alternatives unexplored, while information-seeking actions may incur substantial immediate regret. Information-directed sampling offers a way to balance these competing objectives and can often outperform standard Thompson samples \citep{russo2014ids}. Thanks to efficient estimation of information gain our approach allows to acquire such acquisition samples.

Our method supports two approaches to action selection. First, we can generate multiple approximate Thompson samples and use decision-theoretic criteria to select a desirable candidate. These include expected information gain about the optimal action, expected immediate regret, one-step look-ahead regret, and an information-directed criterion that balances immediate regret against information gain. This approach avoids an additional acquisition-optimization step while allowing more explicit control over exploration and exploitation than selecting a  Thompson sample.

Alternatively, we can optimize the chosen criterion directly over the admissible action space, using the flows to approximate the criterion rather than to generate candidate actions. A finite set of Thompson samples may miss actions that optimize the acquisition criterion. Direct optimization can also identify revealing actions outside the support of the Thompson-sampling distribution: actions that are not plausible reward maximizers under the current belief but provide valuable information about which action is optimal. The flows make these criteria efficient to estimate by providing tractable conditional densities and rapid conditioning on hypothetical observations, although the acquisition criterion computation and optimization introduces additional latency, as compared to amortized Thompson samples which are available immediately, once the data is acquired.

We demonstrate the proposed framework on two model-based design problems from synthetic biology. The first concerns maximizing fluorescent-protein production using a host-aware simulator. The second concerns the design of a genetic oscillator represented by a highly stochastic and non-differentiable model. These applications provide challenging test cases involving intrinsic stochasticity, uncertain experimental conditions, and simulator-based optimization, while illustrating the broader applicability of the framework to scientific design problems.

\paragraph{Contributions.}
Our contributions fall into two groups: amortizing Thompson sampling
and using the resulting models for decision-focused experimental design.

\begin{itemize}
    \item \textbf{Amortized Thompson sampling through parameter inference.}
    We train a parameter-conditioned policy that maps unknown model
    parameters to approximately reward-maximizing actions. Combined
    with amortized simulation-based inference, this approach generates
    approximate Thompson samples without repeated online inference
    or reward optimization.

    \item \textbf{Direct amortization with an action flow.}
    We introduce a conditional normalizing flow that approximates the
    Thompson-sampling distribution directly from experimental histories,
    bypassing explicit parameter inference when generating actions.
    Its tractable conditional density also enables estimation of
    entropy and information gain about the optimal action.

    \item \textbf{Decision-focused selection among Thompson samples.}
    We combine the learned models to efficiently estimate expected
    immediate regret, expected information gain about the optimal
    action, one-step look-ahead regret, and information-directed
    criteria. These criteria enable selection among Thompson-sampled
    candidates, providing explicit control over exploration and
    exploitation beyond posterior randomization.

    \item \textbf{Acquisition optimization beyond Thompson samples.}
    We use the same models to estimate acquisition criteria during
    direct optimization over the admissible action space. This avoids
    restricting decisions to a finite set of Thompson samples and
    allows selection of informative experiments that need not
    themselves be plausible reward maximizers.
\end{itemize}

\subsection{Background and Related Work}

\paragraph{Acquisition functions and sequential decision-making.}
Our work is related to Bayesian optimization (BO), where acquisition functions such as expected improvement \citep{jones1998efficient} and upper confidence bounds (UCB) \citep{srinivas2009gaussian} score candidate actions while balancing exploration-exploitation dilemma.  BO commonly places a probabilistic surrogate, such as a Gaussian process, directly over an unknown objective. Our setting instead assumes access to a simulator model whose behaviour depends on latent parameters and whose likelihood may be intractable.

Related exploration--exploitation strategies arise in the bandit literature, which are also often used for BO \citep{garnett}, such as Thompson sampling. Thompson sampling selects an action by sampling a plausible hypothesis from the current posterior and optimizes the objective under that hypothesis \citep{thompson1933,daniel2018tutorial}. Its exploratory behaviour therefore arises implicitly through posterior randomisation. Information-directed sampling instead explicitly trades off expected regret against information gained about the optimal action \citep{russo2014ids}. Other approaches focus on the information gain alone. Entropy Search \citep{hennig2012entropy} and Predictive Entropy Search
\citep{hernandez2014predictive} select evaluations to gain information
about the location of the maximum of a GP-modelled latent objective, while Max-value
Entropy Search \citep{wang2017max} targets its maximum value. In contrast, we use simulator models without a GP surrogate and amortize
posterior inference and the mapping from latent parameters to optimal actions.
 Acquisition functions such as expected improvement \citep{jones1998efficient} or knowledge gradient \citep{frazier2008knowledge} are based on one-step lookahead approach \citep{garnett}. Inspired by these approaches, one of our selection strategies is the one-step look-ahead regret of a Thompson sample. A further alternative is to formulate the complete adaptive design process as a sequential reinforcement learning problem and learn a policy end-to-end, see e.g. \citet{kobiela2026sequential}.  

\paragraph{Variants of Thompson (posterior) sampling}
Thompson sampling has sublinear cumulative-regret guarantees under established assumptions in a range of bandit settings \citep{daniel2018tutorial}. However, efficient reward accumulation does not imply efficient identification of the optimal action. As \cite{russo2020simple} observes, standard Thompson sampling eventually concentrates measurements on the estimated-best arm, potentially requiring many additional measurements to rule out competing alternatives. Top-two Thompson sampling \citep{russo2020simple} addresses this limitation in finite-armed best-arm identification by selecting a posterior-sampled optimal arm with a prescribed probability and otherwise resampling until a different arm is optimal. In continuous action spaces with a non-atomic distribution over optimal actions, independently sampled Thompson samples are almost surely distinct, so this rejection rule does not identify meaningfully competing alternatives. Our approach enables action selection based on expected information gain about the optimal action, either by ranking Thompson-sampled candidates or by optimizing this criterion directly. This can avoid uninformative measurements and focus experimental effort on identifying the optimal action. 

Thompson sampling was also used in the context of Bayesian reinforcement learning \citep{strens2000bayesian,Osband2013MoreER} where the Markov Decision Process (MDP) is unknown and sampled from the belief state, and afterwards, the policy for that sampled MDP is deployed. In this context it is often known as Posterior Sampling for Reinforcement Learning (PSRL). 
\paragraph{Bayesian optimal experimental design.}
Bayesian optimal experimental design (BOED) selects experiments by maximizing their expected utility, commonly defined as the expected information gain about unknown model parameters \citep{lindley1956measure, chaloner1995bayesian}. Computing this utility typically involves nested expectations over parameters and possible observations, making it expensive to estimate and optimize \citep{foster2019variational,foster2020unified,kleinegesse2019efficient,kleinegesse2020bayesian}. Amortized approaches, including Deep Adaptive Design and related methods, shift this cost to an offline training stage by learning a policy that maps experimental histories directly to subsequent designs \citep{foster2021deep, ivanova2021implicit}. At deployment, the policy can therefore adapt to observations without solving a new optimization problem at every round. We adopt this amortisation principle but focus information acquisition on the downstream objective rather than on identifying the model parameters. In particular, we consider utilities defined in terms of the optimal action $a^\star$ (rather than parameters $\theta$), so that experiments are valued according to downstream task. This objective-focused formulation can avoid acquiring information about parameters that is irrelevant to identifying high-performing actions. A recent approach Action-BED \citep{rossa2026actionbedtaskdrivenbayesianexperimental} avoids using EIG in favor of future loss, often measuring the proximity of the policy output to the true parameters.

\paragraph{Simulation-based inference and normalising flows.}
When a simulator can generate observations but does not provide a tractable likelihood, simulation-based inference (SBI) can be used to perform inference using a machine learning model trained on simulated parameter--observation pairs \citep{thomas2022likelihood, papamakarios2016fast}. Neural posterior estimation uses conditional density estimators, often based on normalising flows \citep{rezende2015variational}, to amortize inference across observations \citep{papamakarios2016fast}. After training, posterior samples and density evaluations can be obtained rapidly for new observations. We learn an amortized posterior over latent parameters and map its samples to optimal actions using a separately trained policy.

\section{Method}
\label{sec:method}

\subsection{Problem formulation}
\label{sec:problem-formulation}

Let $\theta$ denote unknown continuous simulator parameters,
$a\in\mathcal{A}$ a continuous experimental design, and $x$ a discrete
or continuous observation generated by
\begin{align}
x \sim p(x\mid a,\theta).
\end{align}
We assume that the parameter value is fixed throughout an experimental
trajectory and that the conditional reward and observation distributions are
stationary. Given a realised reward $r(a,x,\theta)$, we define
\begin{align}
R(a,\theta)
=
\mathbb{E}_{x\sim p(\cdot\mid a,\theta)}
[r(a,x,\theta)]
\label{eq:expected-reward}
\end{align}
as the expected simulator reward. 
% This expectation accounts for aleatoric uncertainty.

For fixed parameters, an optimal action satisfies
\begin{align}
a^\star(\theta)\in\arg\max_{a\in\mathcal{A}}R(a,\theta).
\label{eq:optimal-action}
\end{align}
We assume a fixed selection rule when the maximum is not unique.
After observing the history
$h_{1:t}=((a_1,x_1),\ldots,(a_t,x_t))$, Thompson sampling draws
\begin{align}
\theta_s &\sim p(\theta\mid h_{1:t}),
&
a_s^\star &\in \arg\max_{a\in\mathcal{A}}R(a,\theta_s).
\label{eq:conventional-ts}
\end{align}
Thus, parameter sampling represents epistemic uncertainty, while
reward optimization averages over aleatoric uncertainty. Conventional
implementations require posterior inference as observations accumulate
and a separate optimization for each posterior sample.

Our framework first amortizes these computations through a learned
policy and conditional density models. It then uses the same models
to select among Thompson-sampled candidates or to optimize
decision-theoretic criteria directly over the action space.

\subsection{Amortized Thompson sampling through parameter inference}
\label{sec:policy-amortisation}

\paragraph{Learning the parameter-to-action mapping.}
We train a parameter-conditioned policy $\pi_\psi(a\mid\theta)$,
parameterised by a feed-forward neural network, to maximize
\begin{align}
J(\psi)
&=
\mathbb{E}_{\theta\sim p(\theta)}
\mathbb{E}_{a\sim\pi_\psi(\cdot\mid\theta)}
[R(a,\theta)]
\nonumber\\
&=
\mathbb{E}_{\theta,a,x}[r(a,x,\theta)].
\label{eq:policy-objective}
\end{align}

We would like to optimize this objective using stochastic gradient descent. However, we assume that the simulator can be an implicit black-box model without available gradients. For this reason, we use the score-function gradient estimator, which takes the form

\begin{align}
\nabla_\psi J(\psi)
=
\mathbb{E}_{\theta,a,x}
\left[
r(a,x,\theta)\nabla_\psi
\log\pi_\psi(a\mid\theta)
\right].
\label{eq:reinforce}
\end{align}
In practice, we use Proximal Policy Optimization (PPO) to stabilise
policy-gradient updates. Training requires derivatives of the policy,
but not of the simulator or reward. It therefore accommodates
stochastic, non-differentiable simulators, including Markov jump
processes.

At deployment, we use the conditional policy mean
\begin{align}
\mu_\psi(\theta)
=
\mathbb{E}_{a\sim\pi_\psi(\cdot\mid\theta)}[a]
\end{align}
as an approximation to $a^\star(\theta)$. 

\paragraph{Amortising parameter inference.}
For each experimental round $t$, we train a conditional normalising
flow
\begin{align}
q_\phi^{(t)}(\theta\mid h_{1:t})
\approx p(\theta\mid h_{1:t})
\end{align}
on simulated parameter--history pairs. Once observations become
available, approximate Thompson samples are generated by
\begin{align}
\theta_s &\sim q_\phi^{(t)}(\cdot\mid h_{1:t}),
&
\widehat a_s^\star &= \mu_\psi(\theta_s).
\label{eq:parameter-policy-ts}
\end{align}
This replaces per-history inference and per-sample reward optimization
with evaluations of pretrained models. Before the first experiment,
parameters are sampled directly from the prior.

\subsection{Direct amortisation with an action flow}
\label{sec:flow-amortisation}

The parameter-flow--policy route generates actions through an explicit
parameter posterior. Our second route learns the distribution over
approximately optimal actions directly:
\begin{align}
q_\omega^{(t)}(\widehat a^\star\mid h_{1:t})
.
\label{eq:action-flow}
\end{align}
To construct training examples, we sample $\theta_i\sim p(\theta)$,
simulate an experimental history under $\theta_i$, as in  Algorithm \ref{alg:flow-training}, and associate that
history with the target
$\widehat a_i^\star=\mu_\psi(\theta_i)$. The action flow is fitted by
conditional maximum likelihood.

At deployment, drawing
\begin{align}
\widehat a_s^\star
\sim q_\omega^{(t)}(\cdot\mid h_{1:t})
\label{eq:direct-ts}
\end{align}
bypasses explicit parameter inference. In addition to sampling, the
flow provides conditional log densities, which are used below to
estimate uncertainty about the optimal action.

Both routes produce approximate Thompson samples.
For action generation, either route can be used. In the present
implementation, parameter flows additionally support regret estimation
and posterior predictive simulation, while action flows provide the
densities required by the information-based criteria.

\paragraph{Training across experimental rounds.}
We train round-specific flows sequentially over a prescribed horizon
of $T$ observations. Histories are generated using an equal mixture
of uniformly sampled actions and approximate Thompson samples.
At the first round, Thompson samples are generated from the prior and
policy, while subsequent rounds use previously trained flows.

For each training trajectory, the ground-truth parameter value is held
fixed. In contrast, parameters used to generate
actions are drawn independently from the belief conditioned on the
current history, to reflect the uncertainty that an experimenter unaware of ground truth would have. The parameter and action flows for a given round are
trained on the same simulated dataset. The full pseudocode can be seen in Algorithm \ref{alg:flow-training}.

\begin{algorithm}[t]
\caption{Sequential training of parameter and action flows}
\label{alg:flow-training}
\KwIn{Prior $p(\theta)$; simulator $p(x\mid a,\theta)$;
policy mean $\mu_\psi$; horizon $T$;
training examples per round $M$}
\KwOut{Flows $q_\phi^{(n)}$ and $q_\omega^{(n)}$
for $n=1,\ldots,T$}

\For{$n=1$ \KwTo $T$}{
    Initialise $\mathcal{D}_n\leftarrow\varnothing$\;
    \For{$i=1$ \KwTo $M$}{
        Sample $\theta_i\sim p(\theta)$\;
        Set $\widehat a_i^\star\leftarrow\mu_\psi(\theta_i)$\;
        Initialise history $h\leftarrow()$\;
        \For{$t=1$ \KwTo $n$}{
            Sample $u\sim\operatorname{Uniform}(0,1)$\;
            \eIf{$u<0.5$}{
                Sample $a_t$ uniformly from $\mathcal{A}$\;
            }{
                \eIf{$t=1$}{
                    Sample $\widetilde\theta\sim p(\theta)$\;
                    Set $a_t\leftarrow\mu_\psi(\widetilde\theta)$\;
                }{
                    Generate $a_t$ using the round-$(t-1)$
                    parameter-flow--policy route or action flow,
                    conditioned on $h$\;
                }
            }
            Sample $x_t\sim p(\cdot\mid a_t,\theta_i)$\;
            Append $(a_t,x_t)$ to $h$\;
        }
        Add $(\theta_i,\widehat a_i^\star,h)$
        to $\mathcal{D}_n$\;
    }
    Fit $q_\phi^{(n)}$ and $q_\omega^{(n)}$
    by conditional maximum likelihood on $\mathcal{D}_n$\;
}
\end{algorithm}

\subsection{Decision-focused selection among Thompson samples}
\label{sec:selection-strategies}

Standard Thompson sampling executes one posterior-sampled optimal
action. Instead, we generate a candidate set
$\mathcal{C}_t=\{a^{(1)},\ldots,a^{(K)}\}$ using either amortized
sampling route and score its members with a decision criterion.
This avoids additional acquisition optimization while allowing
explicit control over exploration and exploitation.

Let $h_{1:t}^+(a)$ denote the history
obtained by appending a hypothetical pair $(a,x)$, where $x$ is sampled from the predictive distribution using current belief. At $t=0$, the
parameter belief is the prior; otherwise, it is approximated by
$q_\phi^{(t)}(\theta\mid h_{1:t})$.

\paragraph{Expected immediate regret.}
We use the policy-based reward reference
\begin{align}
\widehat R^\star(\theta)
=
R(\mu_\psi(\theta),\theta).
\end{align} The PPO value function can generally be used to approximate this quantity.
The expected immediate regret is approximated by
\begin{align}
\widehat{\mathcal{R}}_{\mathrm{now}}(a)
=
\mathbb{E}_{\theta\sim q_\phi^{(t)}(\cdot\mid h_{1:t})}
\left[
\widehat R^\star(\theta)-R(a,\theta)
\right].
\label{eq:immediate-regret}
\end{align}
Minimising this criterion is equivalent to maximizing posterior
expected reward, since the reference does not depend on the candidate
action.

\paragraph{One-step look-ahead regret.}
To assess the usefulness of an experiment for a subsequent decision,
we simulate its outcome and evaluate the expected regret of a
Thompson sample drawn from the updated belief:
\begin{align}
\widehat{\mathcal{R}}_{\mathrm{lookahead}}(a)
={}&
\mathbb{E}_{\theta\sim q_\phi^{(t)}(\cdot\mid h_{1:t})}
\mathbb{E}_{x\sim p(\cdot\mid a,\theta)}
\nonumber\\
&\quad
\mathbb{E}_{\theta'\sim q_\phi^{(t+1)}(\cdot\mid h_{1:t}^+)}
\left[
\widehat R^\star(\theta)
-
R(\mu_\psi(\theta'),\theta)
\right].
\label{eq:future-regret}
\end{align}
This criterion values the candidate through the quality of the
subsequent recommendation, rather than its immediate reward.
An experiment with substantial immediate regret can therefore be
preferred if its observation improves future decisions.

\paragraph{Information gain about the optimal action.}
The EIG about optimal action is defined by:
\begin{align}
\operatorname{EIG}(a)
=
I(a^\star;x\mid h_{1:t},a).
\label{eq:information-gain}
\end{align}
This can be written as
\begin{align}
\operatorname{EIG}(a)
=
H(a^\star\mid h_{1:t})
-
\mathbb{E}_{x\sim p(\cdot\mid h_{1:t},a)}
[H(a^\star\mid h_{1:t}^+)],
\end{align} where entropy is defined as $H(Z)=\mathbb{E}_{Z\sim p}[-\log p(Z)]$.
Thus, it is the expected reduction in uncertainty
about the Thompson-sampling distribution.

The action flow approximates the distribution over policy-generated
optimal actions. Its entropy can be estimated using
\begin{align}
H[q_\omega^{(t)}(\cdot\mid h_{1:t})]
\approx
-\frac{1}{S}\sum_{s=1}^{S}
\log q_\omega^{(t)}(z_s\mid h_{1:t}),
\qquad
z_s\sim q_\omega^{(t)}(\cdot\mid h_{1:t}).
\label{eq:entropy-estimate}
\end{align}
Predictive observations are generated by drawing parameters from the
current belief and simulating $x\sim p(\cdot\mid a,\theta)$.
The next-round action flow then supplies the conditional entropy
for each hypothetical outcome.

\paragraph{Information-directed selection.}
Information-directed sampling combines immediate regret and information gain through
\begin{align}
\widehat{\Psi}(a)
=
\frac{
\widehat{\mathcal{R}}_{\mathrm{now}}(a)^2
}{
\widehat{\operatorname{EIG}}(a)
}.
\label{eq:ids}
\end{align}
We can estimate both of those quantities as before.

To improve comparisons between candidates, we reuse the same
Monte Carlo samples of $\theta $ across all candidate actions.
This common-random-numbers approach can reduce the variance of
estimated loss differences, making candidate rankings more reliable.

\subsection{Candidate selection}
For each minimisation criterion
$L\in\{\widehat{\mathcal{R}}_{\mathrm{now}},
\widehat{\mathcal{R}}_{\mathrm{lookahead}},\widehat{\Psi},-\operatorname{IG}\}$,
the selected action is
\begin{align}
a_{t+1}\in\arg\min_{a\in\mathcal{C}_t}L(a).
\end{align}
Restricting selection to approximate Thompson samples allows us to simply acquire candidates using previously trained flows, but may exclude actions which are informative, but not included in the support of Thompson sampling distribution.

% % seems to be a duplicate of the paragraph at 454-457
% kept the one above since the comment may also apply to the optimisation?
% To improve comparisons between candidates, we reuse the same
% Monte Carlo samples of $ \theta $ across all candidate actions.
% This common-random-numbers approach can reduce the variance of
% estimated loss differences, making candidate rankings more reliable.

\subsection{Direct acquisition optimization}
\label{sec:acquisition-optimization}

Rather than restricting selection to a finite candidate set, we can
optimize the same criteria over the full admissible action space:
\begin{align}
a_{t+1}\in\arg\min_{a\in\mathcal{A}}L(a),
\end{align}
with maximization used for information gain. Here, the flows only estimate
the acquisition criterion rather than also generating the candidate actions.

This approach can recover promising actions missed by a finite
Thompson-sampled set. It can also identify informative experiments
that are not plausible reward maximizers under the current belief
and therefore need not lie in the support of the
Thompson-sampling distribution. 

Further, optimization methods such as Bayesian optimization can potentially use
lower-cost Monte Carlo estimates of these criteria based on fewer
samples, reducing the time required per evaluation, as compared to selection strategies. This is justified as Bayesian
optimization seeks a minimum of the mean
despite noise \citep{garnett}.

\section{Results}
\subsection{Host-aware heterologous gene expression}

In this section we focus on maximizing foreign gene expression using host-aware model. This case study is adopted from \cite{kobiela2026sequential} and its detailed formulation can be found in the appendix. The simulator models the production of green fluorescent protein (GFP) in genetically modified \textit{Escherichia coli}. It represents both GFP production and the host-cell processes that are required to support it, including nutrient uptake, energy generation, transcription, protein synthesis, degradation, and cell growth. The model therefore captures competition between GFP production and normal cellular functions for limited resources, particularly energy and ribosomes, which are the cellular machinery responsible for producing proteins. The model involves 22 coupled ordinary differential equations with nonlinear Hill-type regulatory functions.

\begin{figure}
    \centering
    \includegraphics[width=0.70\linewidth]{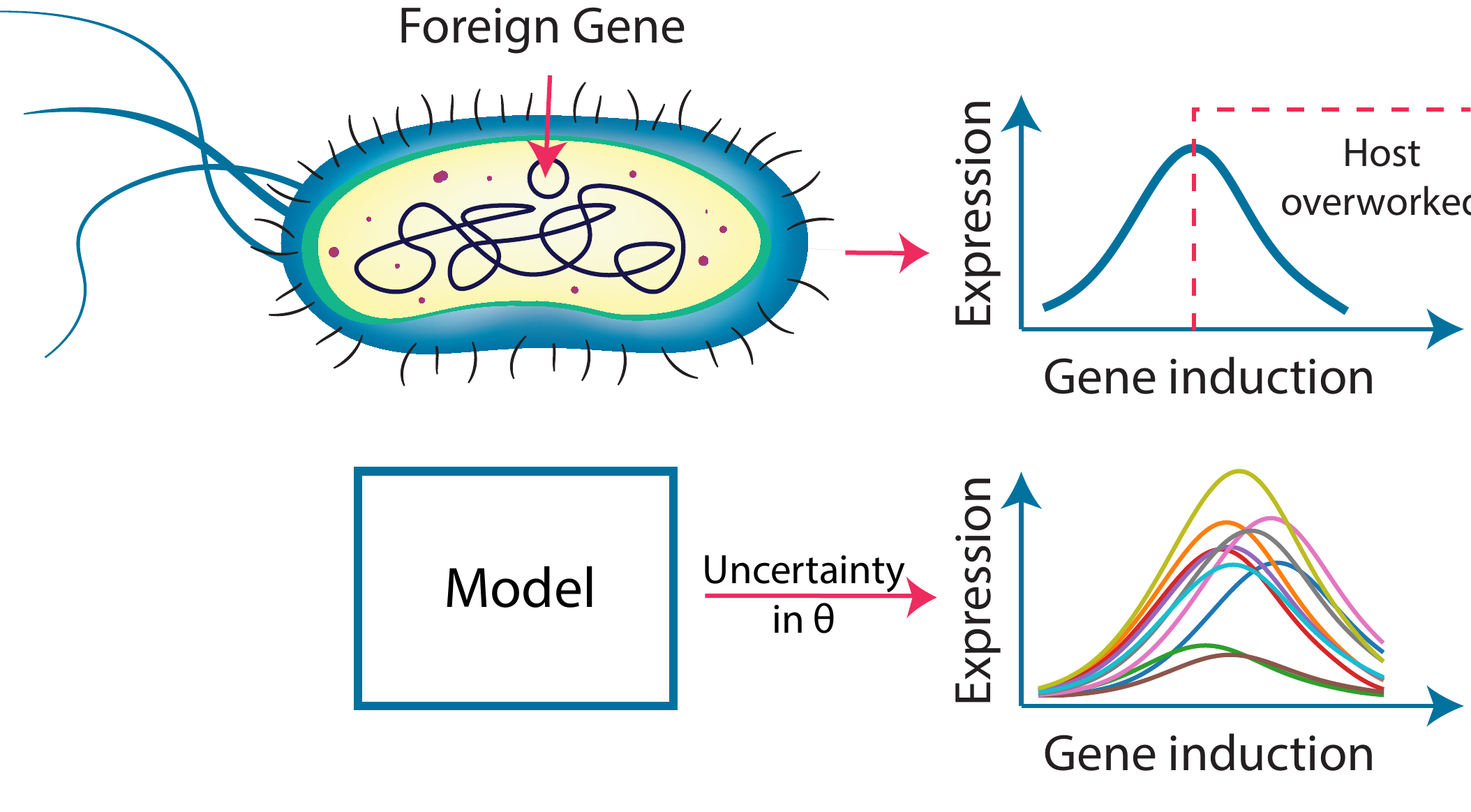}
    \caption{We model foreign gene insertion. The excessive gene induction (designable parameter) can lead to an overworked host. This can decrease the expression of the gene and slow down its growth. The model can be used to predict the behaviour of the system but a lack of knowledge results in uncertainty.}
    \label{fig:overworked}
\end{figure}

The main experimental input is the GFP induction level, $w_{\mathrm{gfp},\max}$, which determines how strongly GFP production is activated. The primary simulator output is the steady-state GFP abundance. This value is obtained by solving the model's system of ordinary differential equations over a sufficiently long time horizon using the \emph{Rosenbrock23} method from the DifferentialEquations.jl library \citep{rackauckas2017differentialequations}. Although increasing induction initially promotes GFP production, excessive induction can overload the host cell and reduce the final protein yield. The resulting relationship between induction and GFP output can therefore be non-monotonic, see Figure \ref{fig:overworked}.

Variation in the biological system is represented by three parameters: the nutrient efficiency $n_s$, the rate at which ribosomes bind to GFP messenger RNA $k_b$, and the rate at which they unbind $k_u$. In principle, each of these parameters can be influenced experimentally, for example through the choice of growth medium or the design of the gene-expression construct. However, their exact values are generally unknown because they are affected by unmodelled regulation, experimental variability, and limited biological knowledge. We therefore treated them as uncertain parameters with uniform prior:
\begin{equation}
n_s \sim \mathcal{U}(0,1), \qquad
k_b \sim \mathcal{U}(0,2), \qquad
k_u \sim \mathcal{U}(0,2).
\end{equation}

Different values of these parameters alter both the maximum achievable GFP output and the induction level at which it occurs, see Figure \ref{fig:overworked}. All remaining model parameters were fixed to the values reported by \citet{nikolados2021prediction}, based on the host-aware model of \citet{weisse2015mechanistic}.  The initial conditions are randomized to represent variation in the starting state of the cells. Random initial conditions induce stochastic observations even though the underlying dynamics are deterministic. Evaluating the likelihood requires integrating over these initial conditions, making the likelihood difficult to evaluate explicitly. This is, however, not an issue as our method does not require explicit likelihoods.

\begin{figure}[h]
    \centering
    \begin{minipage}[c]{0.49\linewidth} \includegraphics[width=\linewidth]{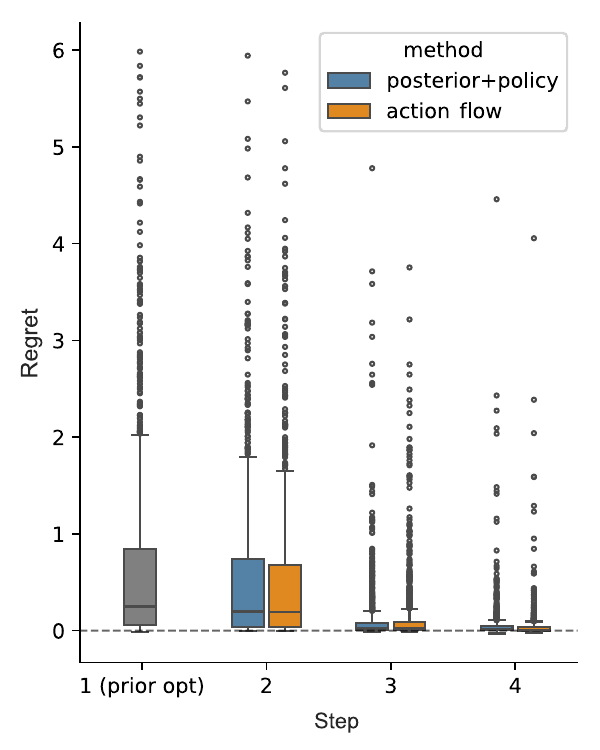}
    \end{minipage}
    \begin{minipage}[c]{0.49\linewidth}
    \includegraphics[width=1\linewidth]{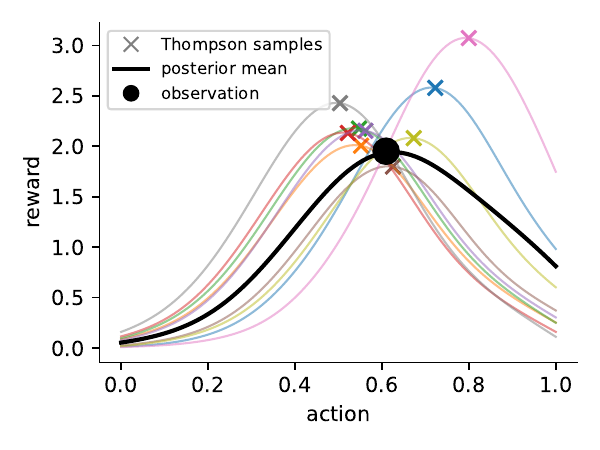}
    \includegraphics[width=1\linewidth]{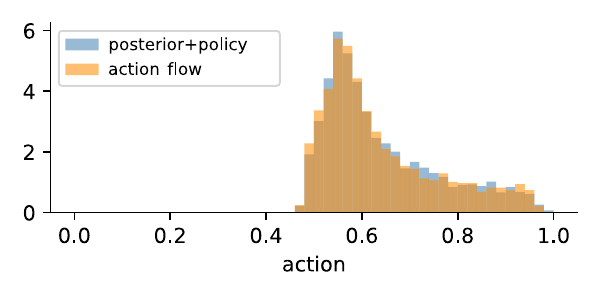}
    \end{minipage}
    \caption{Left: Regret for each step for 1000 rollouts each for different sample of $\theta \sim p(\theta)$. Right: Thompson sample distributions for example context $h_{\text{example}}=(0.61, 1.95)$ using both methods.}
    \label{fig:host-overview}
\end{figure}

We trained both the parameter and action flow for 4 steps. Figure \ref{fig:host-overview} shows the regret over 1,000 rollouts for Thompson sampling, each using a different sample from the prior as the ground truth. It can be seen that, after two steps, the regret improves significantly, with the majority of the probability mass close to zero. However, there is still a large number of outliers. We will show that these outliers can be dramatically reduced using an appropriate Thompson sampling selection or optimization method. 

Figure \ref{fig:host-overview} also shows Thompson samples for the example history context $h_\text{example} = (0.61, 1.95)$. It can be seen that the approximate Thompson samples (crosses) generally maximize the predictive objective function, as desired. Secondly, the Thompson samples generated with both routes---posterior parameter inference followed by the policy, and action-flow---coincide in distribution.

The example context in Figure \ref{fig:host-overview} was chosen because it illustrates several interesting properties. First, a single observation (black dot) is insufficient to reliably identify the optimal action, motivating the need for sequential design. Second, the action that maximizes posterior expected reward almost coincides with the action already observed. Acting purely exploitatively can therefore lead to selecting an action that provides little additional information. Although it has the highest expected reward at the current step, it may contribute little to identifying the true optimum.

\begin{figure}[h]
    \centering
        \includegraphics[width=0.95\linewidth]{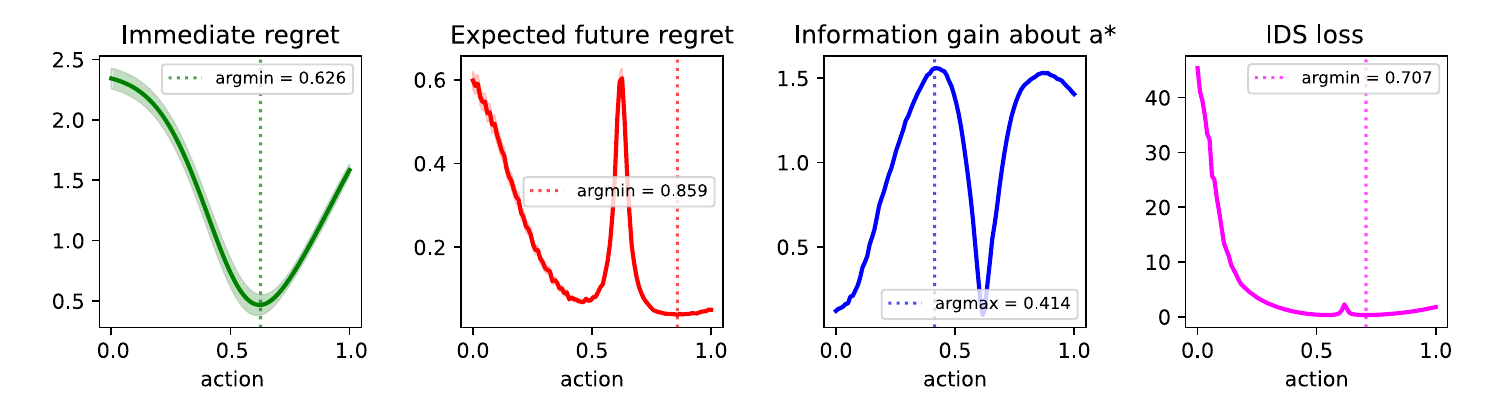}

    \caption{Selection objectives for example history context $h_{\text{example}}=(0.61, 1.95)$.}
    \label{fig:context-select}
\end{figure}

Figure \ref{fig:context-select} shows the selection criteria introduced in Section \ref{sec:selection-strategies} for this history context. The action that minimises expected immediate regret provides little information, as measured by EIG, and all the other criteria favour different, more informative actions. In Appendix Figure \ref{figure:context} and Table \ref{table:context}, we show that selecting such an exploitative action can negatively affect several subsequent decisions, despite yielding a high immediate reward. On the other hand, picking more informative action can yield better results for subsequent steps.

\begin{figure}[t]
    \centering

    \begin{subfigure}{0.89\linewidth}
        \centering
        \includegraphics[width=\linewidth]{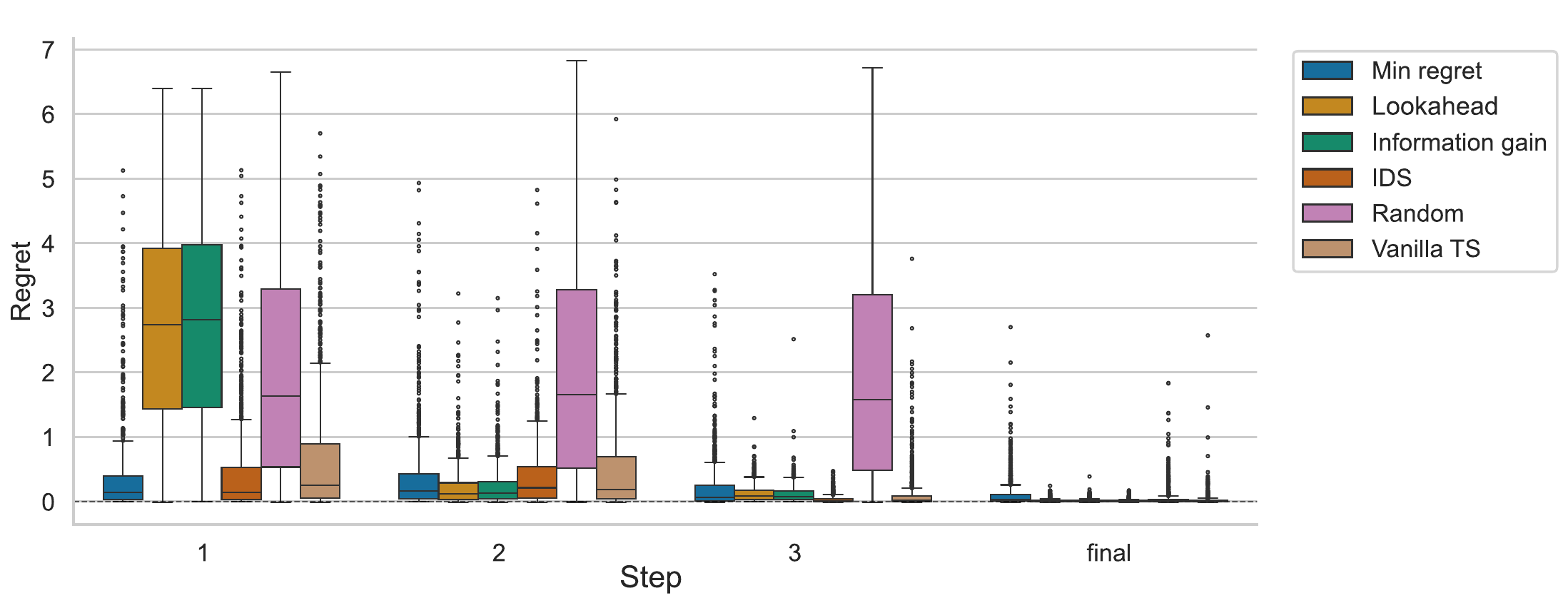}
        \caption{Selection strategies applied to Thompson samples
        over 1,000 rollouts.}
        \label{fig:pooled-selection}
    \end{subfigure}

    \medskip

    \begin{subfigure}{0.89\linewidth}
        \centering
        \includegraphics[width=\linewidth]{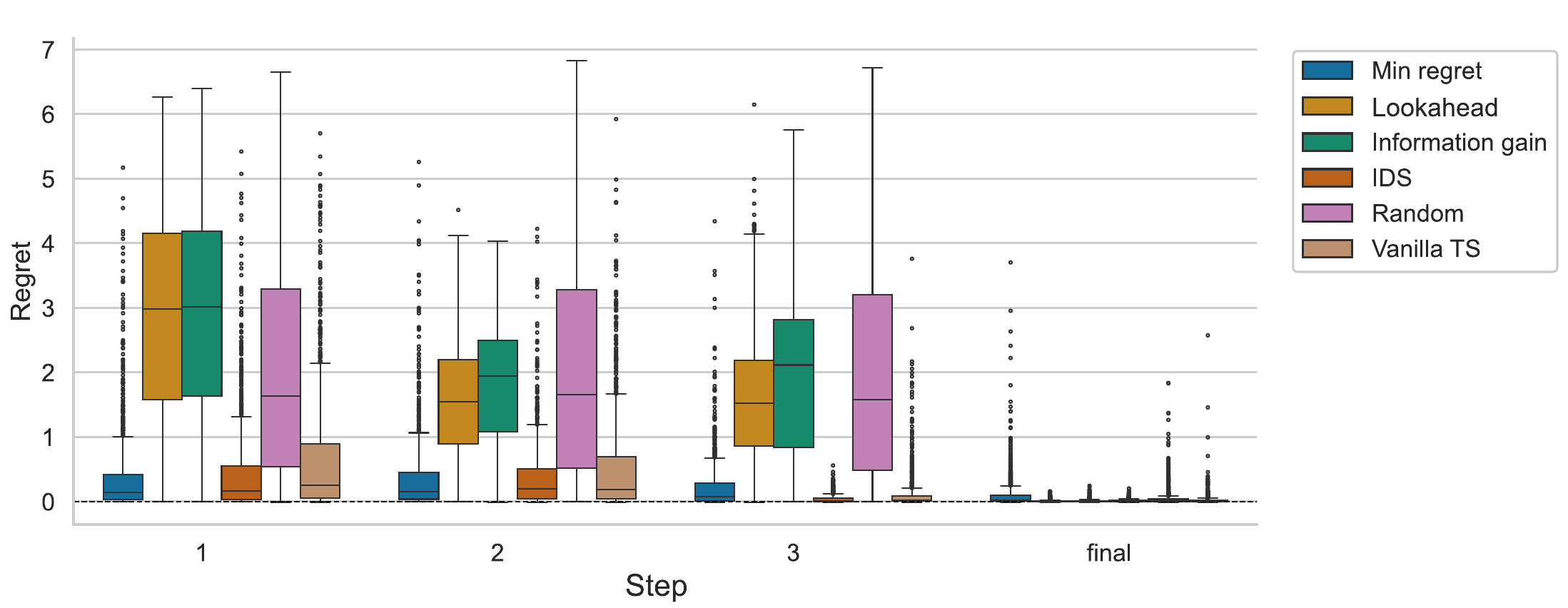}
        \caption{The same strategies optimized using grid search
        over the entire action space.}
        \label{fig:pooled-grid}
    \end{subfigure}

    \caption{Comparison of action-selection strategies.
    \emph{Final} denotes the last step, at which the immediate
    regret strategy was always used.}
    \label{fig:pooled}
\end{figure}

Figure \ref{fig:pooled} shows the results of Thompson-sample selection across the complete experimental trajectory, from the initial to the final step. At the final step, we always use the exploitative strategy based on expected immediate regret, while the strategy of interest is used in the preceding steps. Exploitation at the final step substantially reduces the outlier problem, even when the preceding steps follow vanilla (unfiltered) amortized Thompson sampling. However, using this exploitative selection at every step, rather than only last one, results in the worst final performance as shown in Table \ref{tab:thompson-grid-regret}. Among the Thompson-sample selection strategies, EIG, IDS, and one-step lookahead regret achieved the best final performance. All three significantly outperformed the vanilla Thompson sampling baseline, with no statistically significant differences detected between them. These results highlight that carefully selecting Thompson samples can improve final performance and make experimental design more robust by producing fewer and less severe outliers.

When using optimization rather than selection among Thompson samples, the differences between decision criteria become even more pronounced, as seen in Figure \ref{fig:pooled}b (bottom). One-step lookahead and EIG achieve very low final regret but incur substantial regret in earlier rounds. Searching beyond the Thompson-sampled candidate set allows the selection of more informative actions, but these can incur much higher immediate regret. 

\begin{figure}
    \centering
    \includegraphics[width=0.95\linewidth]{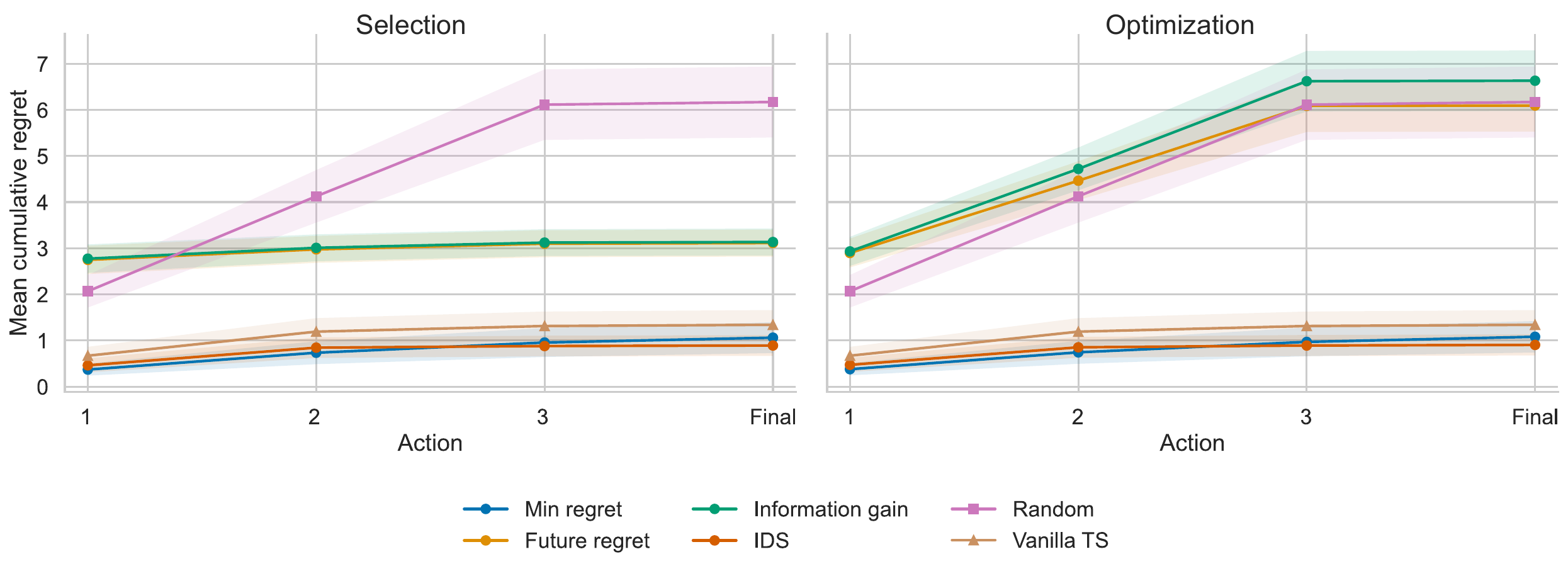}
    \caption{Cumulative regret for 1000 rollouts for GFP case study. The shading indicates standard deviation rescaled by $\frac{1}{5}$ to aid presentation.}
    \label{fig:cummulative-host}
\end{figure}

Figure \ref{fig:cummulative-host} shows that IDS achieved the lowest cumulative regret for Thompson sample selection, whereas EIG and future-regret selection incurred substantial costs in the early rounds, resulting in higher cumulative regret than vanilla Thompson sampling. Whether these strategies are preferable therefore depends on the experimenter's objective. If the priority is to minimise final regret and poorly performing designs can be tolerated during experimentation, EIG and one-step lookahead strategies may be desirable despite their higher cumulative cost. In contrast, IDS remains the strongest strategy when cumulative regret is the primary concern. Notably, it outperforms vanilla Thompson sampling in terms of both cumulative and final regret.

\begin{table}[htbp]
\centering
\small
\caption{Monte Carlo estimates of expected final and cumulative regret,
reported as mean $ \pm $ standard error (SEM). Actions $ a_1 $--$ a_3 $ are
chosen using the specified strategy; the final action is selected by
minimising expected immediate regret. \emph{Italics} identify the
best-performing methods within each selection group (Selection or
Optimization), and \textbf{bold} identifies the best-performing methods
overall. Highlighted methods include the lowest-mean method and those
not significantly different from it under paired $ t $-tests with
Holm correction ($ \alpha = 0.05 $).
}
\label{tab:thompson-grid-regret}
\begin{tabular}{llrr}
\toprule
Approach & Criterion & Final & Cumulative \\
\midrule
Selection & Immediate regret & $ 0.105 \pm 0.007 $ & $ 0.955 \pm 0.050 $ \\
 & One-step lookahead & $ \mathit{0.014} \pm \mathit{0.001} $ & $ 3.098 \pm 0.046 $ \\
 & Information gain & $ \mathit{0.013} \pm \mathit{0.001} $ & $ 3.122 \pm 0.046 $ \\
 & IDS & $ \mathit{0.012} \pm \mathit{0.001} $ & $ \bm{\mathit{0.879}} \pm \bm{\mathit{0.035}} $ \\
\midrule
Optimization & Immediate regret & $ 0.111 \pm 0.009 $ & $ 0.969 \pm 0.049 $ \\
 & One-step lookahead & $ \bm{\mathit{0.007}} \pm \bm{\mathit{0.001}} $ & $ 6.086 \pm 0.089 $ \\
 & Information gain & $ 0.009 \pm 0.001 $ & $ 6.622 \pm 0.104 $ \\
 & IDS & $ 0.013 \pm 0.001 $ & $ \bm{\mathit{0.890}} \pm \bm{\mathit{0.036}} $ \\
\midrule
n/a & Random & $ 0.057 \pm 0.005 $ & $ 6.113 \pm 0.121 $ \\
 & Vanilla TS & $ 0.027 \pm 0.004 $ & $ 1.313 \pm 0.050 $ \\
\bottomrule
\end{tabular}
\end{table}

\subsection{Repressilator}

\begin{figure}[t]
     \centering
\includegraphics[width=0.70\linewidth]{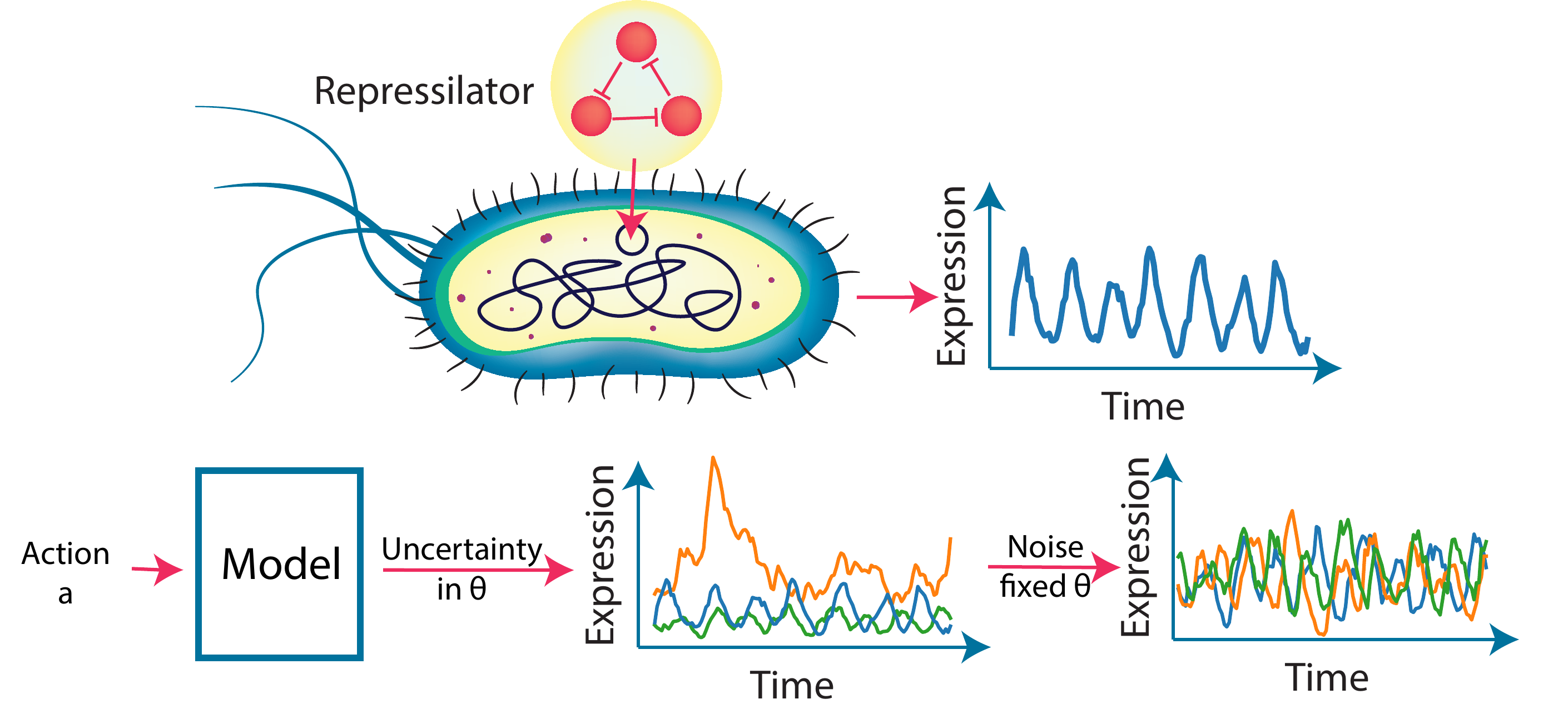}
     \caption{Insertion of three repressing genes (i.e. the \emph{repressilator}) is modelled. Under suitable conditions this results in oscillatory expression. The model not only has uncertain parameters; even with fixed parameters returns highly stochastic prediction of the outcome due to the molecular noise.}
     \label{fig:repressilator}
 \end{figure}

In this case study, we design a \emph{repressilator}, a synthetic genetic oscillator in which three genes repress one another in a cycle \citep{elowitz2000synthetic}, see Figure \ref{fig:repressilator}. We adopt the stochastic simulator of \citet{kobiela2026sequential}, based on \citet{sequeiros2023automated}, and observe the molecule-count trajectory of the first protein, $ p_1(t) $, see modelling details in the appendix.

The action $ a=(k_X,k_m,K) $ specifies the protein translation rate, maximum transcription rate, and repression threshold. We seek actions that produce regular oscillations at a target frequency $ f^\star $, despite uncertainty in the biochemical parameters and intrinsic reaction noise. The reward is
$$
R(a,\theta)
=
-\bigl(f(o)-f^\star\bigr)^2
+\lambda C(\tau_2),
\qquad \lambda=0.3,
$$
where $ f(o) $ is estimated from the second peak of the normalized autocorrelation of $ p_1(t) $, and $ C(\tau_2) $ is the peak height, used as a measure of oscillation regularity.

Our method treats the simulator as a black box, requiring only simulated observations and rewards, without simulator derivatives or access to the likelihood or its parameter score. Further simulator details, action bounds, and parameter priors are provided in the Appendix.

\begin{figure}
    \centering
    \includegraphics[width=0.65\linewidth]{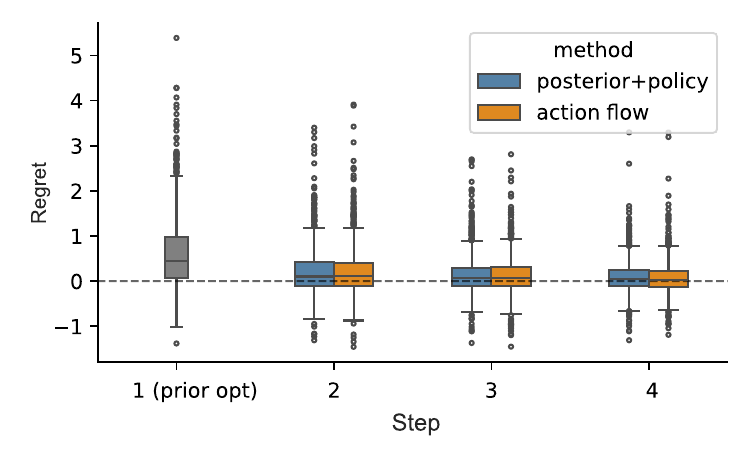}
    \caption{Monte Carlo estimate of the regret across stages for unfiltered Thompson sampling in the repressilator case study.}
    \label{fig:res-repressilator}
\end{figure}

Figure \ref{fig:res-repressilator} presents the results for Thompson samples generated using both amortization methods (summary statistics are provided in Table \ref{tab:regret}). Since rewards are estimated from a highly stochastic simulator and the optimal reward is approximated, estimated regret can be negative. Regret on average decreases across steps, approaching zero. Most of the improvement occurs by the second step, after a single observation has been collected. Some further improvement is observed in the third and fourth step.

\begin{figure}
    \centering
    \includegraphics[width=0.99\linewidth]{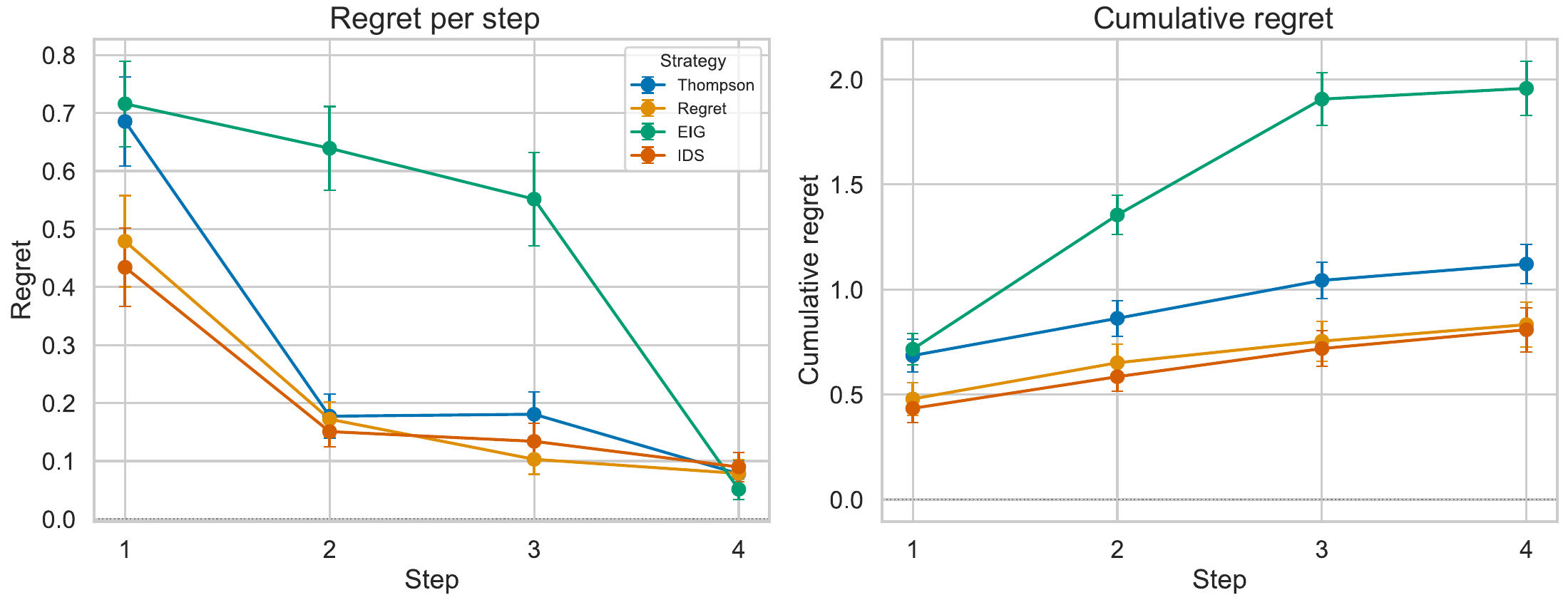}
    \caption{Regret for each acquisition strategy: final regret (left) and cumulative regret (right).}

    \label{fig:rep-full}
\end{figure}

Next, we analyse the optimization of the selection criteria described in Section \ref{sec:acquisition-optimization}. As before, the final action is chosen by pure exploitation. We used 100 ground truth samples from the prior, used expected regret estimated over 100 random seeds to reduce variance. Figure \ref{fig:rep-full} and Table \ref{tab:repressilator-regret} show that EIG achieved the lowest mean final regret, but the other strategies perform similarly in terms of final regret.  Regarding cumulative regret, IDS achieves the best performance, followed by the exploitative immediate-regret strategy, both of them beating Thompson sampling. EIG, on the other hand, incurs substantial regret in earlier rounds.  %This indicates that IDS effectively decides whether exploration is worth the cost when cumulative regret is of interest.

% Required in the preamble:
% \usepackage{booktabs}

\begin{table}[htbp]
\centering
\small
\setlength{\tabcolsep}{5pt}
\caption{Estimated regret for the repressilator case study, reported as
mean $\pm$ standard error.
The specified strategy is used in Steps 1--3; the final action
(Step 4) is selected by minimising expected immediate regret
for all strategies.}
\label{tab:repressilator-regret}
\begin{tabular}{@{}lrrrrr@{}}
\toprule
Criterion & Step 1 & Step 2 & Step 3 & Final & Cumulative \\
\midrule
Thompson sampling
& $ 0.685 \pm 0.077 $
& $ 0.177 \pm 0.038 $
& $ 0.181 \pm 0.038 $
& $ 0.078 \pm 0.024 $
& $ 1.112 \pm 0.072 $\\
Immediate regret
& $ 0.479 \pm 0.079 $
& $ 0.172 \pm 0.030 $
& $ 0.103 \pm 0.026 $
& $ 0.079 \pm 0.023 $
& $ 0.854 \pm 0.075 $\\
Information gain
& $ 0.716 \pm 0.074 $
& $ 0.639 \pm 0.072 $
& $ 0.552 \pm 0.081 $
& $ \bold{0.051 \pm 0.018} $ 
& $ 2.048 \pm 0.081 $ \\
IDS
& $ 0.434 \pm 0.067 $
& $ 0.151 \pm 0.026 $
& $ 0.134 \pm 0.031 $
& $ 0.090 \pm 0.025 $
& $\bold{0.833 \pm 0.075}  $\\
\bottomrule
\end{tabular}
\end{table}

To investigate whether differences between strategies become more pronounced in a more challenging setting, we considered a two-step scenario in which the acquisition strategy is used only once before the final exploitative action. Table \ref{tab:second-case-study-regret} summarises the results. For both selection among Thompson samples and direct optimization, EIG achieved lower mean final regret than the other strategies. Its performance was also comparable to that of vanilla Thompson sampling at the third step in Table \ref{tab:regret}, suggesting that information gain selection may help achieve similar performance with fewer observations.
% Given the differences between mean and median regret, the evaluation across only 100 ground truths, and the averaging of regret for each action over 100 evaluations, we make no claims about statistical significance. 
Overall, the observed patterns are consistent with the previous case study: IDS was preferable when cumulative regret was the primary objective, whereas EIG performed best when the quality of the final recommendation was the main concern.

\begin{table}[htbp]
\centering
\small
\caption{Two-step case study: each strategy selects the first action, followed by an exploitative final action. Final regret is reported as mean $\pm$ SEM.
\label{tab:second-case-study-regret}}
\begin{tabular}{llrr}
\toprule
Approach & Criterion & Final regret \\
\midrule
Selection
 & Immediate regret
 & $ 0.1697 \pm 0.0317 $ \\
 & Information gain
 & $\bold{0.1052 \pm 0.0236}$ \\
 & IDS
 & $ 0.1767 \pm 0.0536 $\\
\midrule
Optimization
 & Immediate regret
 & $ 0.1445 \pm 0.0311 $\\
 & Information gain
 & $\bold{0.1179 \pm 0.0257}$
\\
 & IDS
 & $ 0.1600 \pm 0.0283 $\\
\bottomrule
\end{tabular}
\end{table}

\subsection{Discussion}
Our results demonstrate that amortising Thompson sampling is effective across simulators evaluated in this work. By moving inference and reward optimization to offline training, the framework enables rapid generation of approximate Thompson samples as experimental observations accumulate. The learned flows also make information-based and look-ahead criteria practical to estimate, allowing us to move beyond standard posterior randomisation when selecting experiments.

A useful direction for future work is to enforce consistency between the learned flows. The action flow and the parameter flow are trained separately, although the action distribution should correspond to mapping posterior parameter samples through the pretrained policy. Future training objectives could encourage agreement between these two representations.

Our results suggest that selecting among Thompson-sampled candidates using explicit decision criteria can improve the reward of the final recommendation relative to unfiltered Thompson sampling. This distinction matters when the objective is to identify a high-performing design after a limited experimental budget, rather than maximize reward accumulated during experimentation. Informative actions early in the experiment can reduce uncertainty about the optimal action and improve subsequent recommendations. Conversely, final round benefits from exploitation with immediate regret criterion. IDS remains the strongest strategy when cumulative regret is of interest.

% In the settings considered here, information-directed selection appears to offer a useful compromise, improving the cumulative regret as compared to standard Thompson sampling. By balancing expected immediate regret against information gain about the optimal action, it can favour informative experiments without disregarding their immediate cost. 

The choice between selecting among Thompson-sampled candidates and directly optimizing an acquisition criterion introduces a further trade-off. A finite candidate set may miss actions that optimize the criterion, while direct optimization can search the full admissible action space. It can therefore identify revealing actions that are never reward-optimal under the current belief but are informative about which action is optimal. Such actions may incur substantial immediate regret, particularly when selected solely for information gain. Restricting selection to Thompson-sampled candidates provides an implicit preference for plausible reward maximizers and can make information-based selection more exploitative, at the cost of excluding most informative experiments.
% This restriction does not guarantee low immediate regret, however, because an action that is optimal under one posterior hypothesis may perform poorly under others.

\section{Conclusion}
We introduced an approach to amortizing Thompson sampling that shifts inference and reward optimization costs ahead of data acquisition. It accommodates simulator models with intractable likelihoods and unavailable gradients, including discrete-state Markov jump processes, and remained feasible for host-aware protein expression models with over 20 coupled ordinary differential equations. Both routes, parameter inference coupled with a pretrained policy and direct sampling from an action flow, were effective in our case studies.

Beyond efficient sampling, we used the learned models to guide experimental design in two ways: selecting among Thompson-sampled candidates using explicit decision criteria, and directly optimizing these criteria over the admissible action space. The action flow enables estimation of information gain about the optimal action, supporting decision-focused Thompson samples selection and direct acquisition optimization. Information-gain-based approach improved final regret relative to unfiltered Thompson sampling, while information-directed sampling improved cumulative regret in both case studies. In the first case study, direct optimization of information-gain and look-ahead criteria further improved final regret as compared to selection, but at the cost of greater regret during earlier experimentation. Our approach therefore allows for more tailored decision-making, accounting for the tolerance for failures during experimentation and the ultimate goal: maximizing final reward or minimizing cumulative regret.

Our approach reduces computational delays between data acquisitions with amortization and supports more efficient use of experimental budgets with decision criteria. It accommodates broad range of models including those with intractable likelihoods and unavailable gradients.  Together, these advances enable more efficient design using simulator models, which are widely used in science and engineering.

\paragraph{Acknowledgements: }

This work was supported by the United Kingdom Research and Innovation (grant EP/S02431X/1), UKRI Centre for Doctoral Training in Biomedical AI at the University of Edinburgh, School of Informatics. For the purpose of open access, the authors have applied a creative commons attribution (CC BY) licence to any author accepted manuscript version arising.

During the preparation of this work the authors used ChatGPT in order to increase the readability of the text. After using this tool/service, the authors reviewed and edited the content as needed and take full responsibility for the content of the publication. The authors also used Claude AI (Opus 4.8) to aid the implementation of the method.

\bibliography{iclr2025_conference}
\bibliographystyle{iclr2025_conference}

\appendix

% Reset table and figure counters
\setcounter{table}{0}
\setcounter{figure}{0}

% Use A1, A2, ... for appendix tables and figures
\renewcommand{\thetable}{A\arabic{table}}
\renewcommand{\thefigure}{A\arabic{figure}}

\renewcommand{\theHtable}{appendix.A.\arabic{table}}
\renewcommand{\theHfigure}{appendix.A.\arabic{figure}}

\section{Appendix}

\subsection{Host-aware model}
This model is adopted from \cite{weisse2015mechanistic}, and the setup is identical to that in \cite{kobiela2026sequential}. The chemical reaction network is as follows:
\begin{equation}
\begin{aligned}
\text{Import:}\quad
& s \xrightarrow{v_{\text{imp}}} s_{\text{int}} \\
\text{Catabolism:}\quad
& s_{\text{int}} \xrightarrow{v_{\text{cat}}} n_s e \\
\text{Transcription:}\quad
& \varnothing \xrightarrow{w_x} m_x \\
\text{Ribosome binding:}\quad
& p_r + m_x
  \overset{k_{bx}}{\underset{k_{ux}}{\rightleftharpoons}} c_x \\
\text{mRNA degradation:}\quad
& m_x \xrightarrow{d_{m,x}} \varnothing \\
\text{Protein degradation:}\quad
& p_x \xrightarrow{d_x} \varnothing \\
\text{Translation:}\quad
& n_x e + c_x \xrightarrow{v_x} p_r + c_x + p_x \\
\text{Energy dilution:}\quad
& e \xrightarrow{\lambda} \varnothing \\
\text{mRNA dilution:}\quad
& m_x \xrightarrow{\lambda} \varnothing \\
\text{Complex dilution:}\quad
& c_x \xrightarrow{\lambda} \varnothing \\
\text{Protein dilution:}\quad
& p_x \xrightarrow{\lambda} \varnothing
\end{aligned}
\label{eq:host-reactions}
\end{equation}

for each protein $x \in \{r, t, m, q, \text{gfp}\}$.

The rates are defined:
\begin{align}
v_{\text{imp}} &= \frac{p_t v_t s}{K_t + s}, \\
v_{\text{cat}} &= \frac{p_m v_m s_i}{K_m + s_i}, \\
w_x &= \frac{w_{x,\max} \, e}{\theta_x + e},
\quad x \in \{r,t,m,\text{gfp}\}, \\
w_q &= w_{q,\max} \frac{e}{\theta_q + e}
\frac{1}{1 + (\rho_q / K_q)^{h_q}}, \\
v_x &= \frac{c_x}{n_x}
\frac{\gamma_{\max} e}{e + K_\gamma}, \\
\lambda &= \frac{\gamma(e)}{M}
\sum_{x \in \{r,t,m,q,\text{gfp}\}} c_x,
\label{lambda} \\
M &= \sum_x n_x p_x + \sum_x n_r c_x, \\
\gamma(e) &= \frac{\gamma_{\max} e}{e + K_\gamma}.
\end{align}
The corresponding system of ODEs was derived and solved using Rosenbrock solver from \cite{rackauckas2017differentialequations}. The initial condition is set by solving a warm-up problem without GFP, and employing the steady state values with additional Gaussian noise.  We used the default values from \cite{weisse2015mechanistic}, except below, for uncertain and design parameters (actions).

Uncertain parameters priors:  $n_s \sim \mathcal{U}(0,1)$, $k_b \sim \mathcal{U}(0,2)$, and $k_u \sim \mathcal{U}(0,2)$ 

Design parameter admissible ranges: $w_{\text{gfp,max}} \in (1,10000)$.

\subsection{Repressilator model}
This model is adopted from \cite{sequeiros2023automated}, and the setup is identical as in \cite{kobiela2026sequential}.
For each gene $ i \in \{1,2,3\} $, the reactions are
\begin{equation}
\begin{aligned}
\text{Transcription:}\quad
& \emptyset \xrightarrow{\alpha_i} m_i \\
\text{Translation:}\quad
& m_i \xrightarrow{k_X} m_i + p_i \\
\text{mRNA degradation:}\quad
& m_i \xrightarrow{\gamma_m} \emptyset \\
\text{Protein degradation:}\quad
& p_i \xrightarrow{\gamma_X} \emptyset
\end{aligned}
\end{equation}
where
\begin{align}
\alpha_1 &= k_m \left(
\epsilon + \frac{1-\epsilon}{1+(p_3/K)^H}
\right), \\
\alpha_2 &= k_m \left(
\epsilon + \frac{1-\epsilon}{1+(p_1/K)^H}
\right), \\
\alpha_3 &= k_m \left(
\epsilon + \frac{1-\epsilon}{1+(p_2/K)^H}
\right).
\end{align}

Catalyst.jl \cite{CatalystPLOSCompBio2023} was used  to derive the MJP problem. Then stochastic simulation algorithm (SSA) from \emph{JumpProcesses.jl} \cite{rackauckas2017differentialequations} was used to simulate the problem.

Uncertain parameters priors: $H \sim \mathcal{U}(3,7)$, $\gamma_X \sim \mathcal{U}(0.8,1.1)$, $\gamma_m \sim \mathcal{U}(4,50)$, and $\epsilon \sim \mathcal{U}(0.05,0.15)$.

Design parameter admissible ranges: $k_X \in (100, 1000), \quad
k_m \in (3, 120), \quad
K \in (10, 200)$.

\clearpage

\subsection{Rollouts for example context}
\label{app:context}
\begin{table}[h]
\centering
\begin{tabular}{llcc}
\hline
Step & Criterion & Mean $\pm$ SEM & Median \\
\hline
\multirow{6}{*}{a2 (selection on)}
 & Information gain      & $1.190 \pm 0.041$ & $0.685$ \\
 & One-step lookahead  & $1.094 \pm 0.018$ & $1.155$ \\
 & IDS     & $0.646 \pm 0.020$ & $0.527$ \\
 & Immediate regret      & $0.540 \pm 0.029$ & $0.222$ \\
 & vanilla      & $0.676 \pm 0.031$ & $0.281$ \\
\hline
\multirow{6}{*}{a3 (selection off)}
 & Information gain     & $0.091 \pm 0.008$ & $0.011$ \\
 & One-step lookahead  & $\mathbf{0.043 \pm 0.003}$ & $0.015$ \\
 & IDS    & $0.108 \pm 0.009$ & $0.031$ \\
 & Immediate regret    & $0.662 \pm 0.030$ & $0.283$ \\
 & vanilla      & $0.193 \pm 0.015$ & $0.036$ \\
\hline
\multirow{6}{*}{final (selection off)}
 & Information gain     & $\mathbf{0.029 \pm 0.003}$ & $0.006$ \\
 & One-step lookahead & $ \mathbf{0.031 \pm 0.002}$ & $0.011$ \\
 & IDS    & $0.049 \pm 0.005$ & $0.014$ \\
  & Immediate regret     & $0.241 \pm 0.018$ & $0.049$ \\
 & vanilla      & $0.088 \pm 0.009$ & $0.018$ \\
\hline
\end{tabular}
\caption{Regret by step and strategy (mean $\pm$ SEM and median) for example context.}
\label{table:context}
\end{table}

\begin{figure}[h]
\centering
\includegraphics[width=0.6\linewidth]{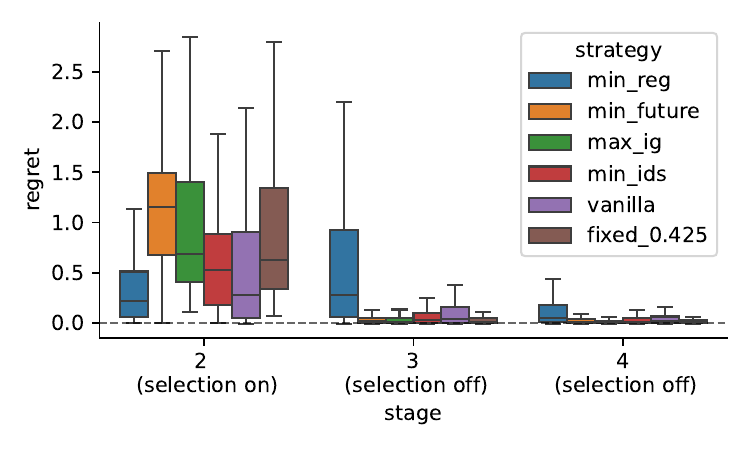}
\caption{Plot presenting the results in Table \ref{table:context}.}
\label{figure:context}
\end{figure}

\subsection{Host study - complete results}

\begin{sidewaystable}[htbp]
\centering
\small
\caption{Monte Carlo estimates of expected regret across trajectory stages,
reported as mean $\pm$ standard error (SEM). Actions $a_1$--$a_3$ are
chosen using the specified strategy; the final action is selected by
minimising expected immediate regret. For final and cumulative regret,
\emph{italics} identify the best-performing methods within each selection
group (Selection or Optimization), and \textbf{bold} identifies the best-performing
methods overall. Highlighted methods include the lowest-mean method and
those not significantly different from it under paired $t$-tests with
Holm correction ($\alpha = 0.05$).
}
\label{tab:thompson-grid-regret-full}
\centering
\begin{tabular}{llrrrrr}
\toprule
Approach & Criterion & Step 1 & Step 2 & Step 3 & Final & Cumulative \\
\midrule
Selection& Immediate regret & $ 0.369 \pm 0.021 $ & $ 0.365 \pm 0.019 $ & $ 0.221 \pm 0.013 $ & $ 0.105 \pm 0.007 $ & $ 0.955 \pm 0.050 $ \\
 & One-step lookahead & $ 2.747 \pm 0.049 $ & $ 0.227 \pm 0.011 $ & $ 0.124 \pm 0.004 $ & $ \mathit{0.014} \pm \mathit{0.001} $ & $ 3.098 \pm 0.046 $ \\
 & Information gain & $ 2.773 \pm 0.049 $ & $ 0.235 \pm 0.010 $ & $ 0.114 \pm 0.004 $ & $ \mathit{0.013} \pm \mathit{0.001} $ & $ 3.122 \pm 0.046 $ \\
 & IDS & $ 0.460 \pm 0.024 $ & $ 0.384 \pm 0.016 $ & $ 0.035 \pm 0.002 $ & $ \mathit{0.012} \pm \mathit{0.001} $ & $ \bm{\mathit{0.879}} \pm \bm{\mathit{0.035}} $ \\
\midrule
Optimization & Immediate regret & $ 0.377 \pm 0.021 $ & $ 0.366 \pm 0.019 $ & $ 0.226 \pm 0.013 $ & $ 0.111 \pm 0.009 $ & $ 0.969 \pm 0.049 $ \\
 & One-step lookahead & $ 2.897 \pm 0.050 $ & $ 1.567 \pm 0.028 $ & $ 1.622 \pm 0.032 $ & $ \bm{\mathit{0.007}} \pm \bm{\mathit{0.001}} $ & $ 6.086 \pm 0.089 $ \\
 & Information gain & $ 2.931 \pm 0.050 $ & $ 1.789 \pm 0.029 $ & $ 1.902 \pm 0.037 $ & $ 0.009 \pm 0.001 $ & $ 6.622 \pm 0.104 $ \\
 & IDS & $ 0.471 \pm 0.024 $ & $ 0.381 \pm 0.017 $ & $ 0.038 \pm 0.002 $ & $ 0.013 \pm 0.001 $ & $ \bm{\mathit{0.890}} \pm \bm{\mathit{0.036}} $ \\
\midrule
n/a & Random & $ 2.067 \pm 0.056 $ & $ 2.059 \pm 0.056 $ & $ 1.988 \pm 0.054 $ & $ 0.057 \pm 0.005 $ & $ 6.113 \pm 0.121 $ \\
 & Vanilla TS & $ 0.669 \pm 0.030 $ & $ 0.522 \pm 0.025 $ & $ 0.122 \pm 0.010 $ & $ 0.027 \pm 0.004 $ & $ 1.313 \pm 0.050 $ \\
\bottomrule
\end{tabular}
\end{sidewaystable}
    
\clearpage

\subsection{Repressilator - additional results}
\begin{table}[h!]
    \centering
        \caption{\label{tab:regret}Estimated (MC) expected regret across four stages,
    with standard error of the mean (SEM) and median regret.}
    \begin{tabular}{lcc}
        \toprule
        Step & Mean $ \pm $ SEM & Median \\
        \midrule
        First action (prior optimization) & $ 0.651 \pm 0.028 $ & $ 0.416 $ \\
        2nd (posterior flow) & $ 0.211 \pm 0.016 $ & $ 0.090 $ \\
        2nd (action flow)    & $ 0.259 \pm 0.018 $ & $ 0.122 $ \\
        3rd (posterior flow) & $ 0.108 \pm 0.013 $ & $ 0.055 $ \\
        3rd (action flow)    & $ 0.136 \pm 0.014 $ & $ 0.061 $ \\
        4th (posterior flow) & $ 0.101 \pm 0.012 $ & $ 0.060 $ \\
        4th (action flow)    & $ 0.083 \pm 0.012 $ & $ 0.051 $ \\
        \bottomrule
    \end{tabular}

\end{table}

\end{document}

%% file: math_commands.tex
\usepackage{amsmath,amsfonts,bm}

\def\eqref#1{equation~\ref{#1}}
\def\1{\bm{1}}

\DeclareMathAlphabet{\mathsfit}{\encodingdefault}{\sfdefault}{m}{sl}
\SetMathAlphabet{\mathsfit}{bold}{\encodingdefault}{\sfdefault}{bx}{n}